\documentclass[11pt,a4paper]{article}
\usepackage{graphicx}
\usepackage[affil-it]{authblk}
\usepackage{grffile}
\usepackage{cite}

\newcommand{\quotes}[1]{``#1''}

\begin{document}
\date{}
\title{Visualizing Natural Language Descriptions: A Survey}
\author{Kaveh Hassani%
\thanks{email: \texttt{kaveh.hassani@uottawa.ca}; Corresponding author}  and Won-Sook Lee%
\thanks{email: \texttt{wslee@uottawa.ca}\\
The complete paper appears in ACM Computing Surveys\\
DOI: http://dx.doi.org/10.1145/2932710 }}
  
\affil{School of Computer Science and Electrical Engineering\\ University of Ottawa, Canada}
\maketitle


\begin{abstract}
\noindent
A natural language interface exploits the conceptual simplicity and naturalness of the language to create a high-level user-friendly communication channel between humans and machines. One of the promising applications of such interfaces is generating visual interpretations of semantic content of a given natural language that can be then visualized either as a static scene or a dynamic animation. This survey discusses requirements and challenges of developing such systems and reports 26 graphical systems that exploit natural language interfaces and addresses both artificial intelligence and visualization aspects. This work serves as a frame of reference to researchers and to enable further advances in the field.
\end{abstract}

\section{Introduction}
Imagination is an irrefutable part of mankind's social-cognitive processes such as visual-spatial skills, memory access, learning, creativity, and communication. People rely on the capability of creating, sharing, and communicating imaginations in their daily activities. There are two general approaches to share imaginations: explicit approach and implicit approach. In former, imaginations are directly realized using visualization techniques such as sketching, painting, and computer-aided tools. This approach has an objective nature and results in an accurate description of a given imagination. However, it requires high-level visualization skills that would take years of practice and learning. As an example, in case of computer animation, a vast variety of design tools are available. These tools provide the designers with a user-friendly graphical user interface (GUI) that follows a dominant approach in human-computer interactions known as windows, icons, menus, and pointer (WIMP) model. Learning these tools even for professional designers is tedious, labor intensive, and time-consuming, requiring them to learn and utilize a set of complex graphical interfaces. Furthermore, migrating from one specific tool to another one would require learning a set of new interfaces from scratch. The learning process faces more sophistication in case of scripting interfaces such as graphical APIs and game engines. The learning curve is even slower for novice users who only need to create a simple animation for ad hoc applications.

The implicit approach, on the other hand, is subjective and is carried out by representing the imaginations through a set of natural language descriptions. In this approach, a speaker or a writer shares her imaginations through a verbal channel and an audience perceives and reconstructs the imaginations based on their internal mental states. Hence, a unique imagination can result in different realizations. Due to simplicity and  naturalness of describing the imaginations through the verbal channel, the explicit approach is considered the dominant mechanism for sharing the imaginations in interpersonal communications. It motivates researchers to develop systems that can directly convert the natural language descriptions to target visualizations.

A natural language interface exploits the conceptual simplicity and naturalness of the language to create a high-level user-friendly communication channel between humans and machine. The interface can be used to generate visual interpretations of the semantic content of a given natural language that can be then visualized either as a static scene or a dynamic animation. Nevertheless, current technical difficulties do not allow machines to completely capture the deep semantics embedded within natural languages. These difficulties root in characteristics of natural languages such as being semi-structured, ambiguous, context-sensitive and subjective.

In recent research \cite{lee2014potential}, comprehensive user studies are performed to compare the performance of natural language interfaces against conventional GUI for animation design tasks in terms of control, creativity, and learning measures. The results suggest that in terms of high-level control over virtual objects and animation design, natural language interfaces outperform GUIs, whereas in terms of spatial and motion control it is simpler to use GUIs. It is also concluded that using GUIs increases the creativity in micro-level design while in macro-level design natural language interfaces are more efficient because of their higher versatility. It is also shown that natural language interfaces significantly reduce both learning time and design time. According to this study, a good strategy is to develop a hybrid interface that integrates both interfaces and lets the user decide which one to use for different manipulations.

\subsection{Requirements and Challenges}
Three general requirements can be identified for developing such systems. The first requirement is associated with computer graphics. A system for visualizing natural language descriptions  requires a visualization engine to realize the final interpretation of the language. Fortunately, the current software and hardware technologies of the computer graphics are highly advanced and can generate natural visualizations in real time. Thus, this requirement does not pose any challenges. The second requirement is related to understanding the natural language. A natural language interface must be able to disambiguate a description, discover the hidden semantics within it, and convert them into a formal knowledge representation. This requirement, even for a limited system, can present a fundamental challenge. The third requirement is designing an integrated architecture. Designing a system capable of integrating a natural language interface and a GUI for visualization purposes requires tackling profound technical challenges in different conceptual and operational levels. Such a system requires integration of artificial intelligence (AI) techniques such as natural language understanding (NLU), knowledge representation (KR), planning, spatio-temporal reasoning, and so on, and computer graphics techniques such as real-time rendering, action synchronization, behavior-based modeling, deformation, and so on, in a consistent manner. Considering the above-mentioned requirements, five main challenges of developing these systems can be identified as follows.

\subsubsection{Natural Language Understanding}
NLU is the process of disambiguating a set of descriptions expressed in natural language, capturing deep semantics embedded within surface syntax, and converting the discovered semantics into a representation that can be processed by software. This process relies on a hierarchy of some sub-processes including but not limited to (1) morphological analysis such as stemming and lemmatization; (2) syntactic analysis such as part-of-speech (POS) tagging, syntactic parsing, named-entities recognition, and anaphora resolution; (3) semantic analysis such as word disambiguation, capturing predicate-argument structures, and role labeling; and (4) discourse analysis. A natural language interface with the visualization purpose should disambiguate the descriptions based on scene arrangements and capture the semantics associated with scene layout, spatio-temporal constraints, parameterized actions, and so on.

\subsubsection{Inferring Implicit Knowledge}
When people communicate, they assume the target audience have a priori knowledge about the context and hence do not elaborate on it. They also omit the common-sense facts and assume the audience fill in the gaps \cite{chang2014interactive}. Inferring this implicit knowledge is a big challenge for current computer software. Furthermore, it is a challenging task to derive meaningful interpretations of spatio-temporal relations from descriptions of the world model.

\subsubsection{Knowledge Representation}
KR refers to a formal representation of information in a way that computer software can utilize it to perform complex tasks. The representation should support insertion, update, and querying operations on the target knowledge. It should also represent concepts, entities, relations, constraints, uncertainty, etc. A system designed to convert natural language descriptions to a visual representation requires a KR component to represent the discovered semantics and use it to decide the actions to be taken. Also, a reasoning mechanism embedded within the KR component can help the system to derive implicit knowledge from available knowledge. Designing such a component is not a trivial task.

\subsubsection{Symbol Grounding}
Semantics are represented as high-level concepts within the KR component that eventually need to be grounded into low-level graphical objects, visual features, transformations, and relations. This mapping process involves decomposing high-level concepts into a set of low-level graphical instructions running in serial or parallel and parametrizing those instructions.  Automating this process is one of the AI research goals.

\subsubsection{Scalability}
A scalable system should couple high-level semantic processing with low-level action decomposition in a consistent manner. It should also exploit data-driven techniques to generalize to unseen scenarios. Gathering required tools and repositories such as lexical resources and object database is also a challenging task. In addition, obtaining the knowledge itself (i.e., both implicit and explicit) is a challenge.

\subsection{Classification}
Considering the interdisciplinary nature of visualization systems with natural language interfaces, one can categorize the literature from several points of view. In terms of design methodology, the systems can be classified into rule-based, data-driven, and multi-agent systems. Another possible classification can be based on the system behavior that divides the systems into reactive and deliberative systems. It is also possible to classify the literature based on the utilized language understanding approach, syntactic analysis, knowledgebase scheme, and so on. However, none of these classifications address the graphical aspects. Similarly, one may categorize the literature based on the graphical aspects that ignore the intelligent aspects of the systems. In order to have a consistent classification scheme that can address different aspects of the research works, we categorize the literature based on the generated output. Using this scheme, we classify the literature to three categories: text-to-picture, text-to-scene, and text-to-animation conversion systems.

It is noteworthy that throughout the article, the term text interchangeably refers to any oral or written form of the language utterance. In this regard, verbal commands issued by an operator, textual scripts provided by a user, or textual content within Web
pages are all treated as text.

\subsection{Contribution}
As far as the authors' knowledge is concerned, this article is the first comprehensive overview on the systems and approaches associated with visualizing natural language descriptions. Surprisingly, despite its scientific and industrial merit, not so many studies have been carried out in this direction. And among existing works, there are only a few that have solid contributions to this field. This survey discusses requirements and
challenges for developing such systems and reports 26 graphical systems that exploit natural language interfaces and addresses both artificial intelligence and visualization aspects. This work serves as a frame of reference to researchers and to enable further advances in the field. For each introduced system, we elaborate on the system inputs and outputs, design methodology, architecture, implementation, language processes, graphical processes, intelligent processes, and resources and discuss the advantages and disadvantages as well.

\subsection{Organization}
The article is organized as follows. Section 2 provides a concise terminology of computational linguistics. Section 3 overviews the text-to-picture conversion systems and investigates two example systems. Section 4 discusses the text-to-scene conversion systems and elaborates on seven systems. Section 5 provides a comprehensive overview of 17 text-to-animation conversion systems. Section 6 discusses the overall restrictions
of the developed systems and provides potential solutions and possible directions for future studies. Section 7 concludes the article.

\section{Terminology of Computational Linguistics}
Considering the interdisciplinary nature of this article, it will possibly attract audiences from different fields such as computational linguistics, human-computer interactions, artificial intelligence, and computer graphics. To provide the readers with a self-contained article, this section provides a concise terminology of computational linguistics as follows.
\begin{itemize}
\item \emph{Stop-Words:} words with syntactic functionality that carry insignificant semantic information (e.g., \emph{the} and \emph{is}).
\item \emph{Bag-of-Words Model:} a text representation model that treats a given text as a set of words and frequencies and disregards the syntax and word order.
\item \emph{Lemmatization:} the process of grouping together the different inflected forms of a word so they can be analyzed as a single item. As an example, \emph{go} is the lemma of the words [go, goes, going, went, gone].
\item \emph{Named-Entity Recognition:} the process of locating and classifying elements in text into pre-defined categories such as the names of persons, organizations, locations, and so on.
\item \emph{POS–Taggingn:} also known as word-category disambiguation; the process of labeling each and every word in a given text by its grammatical category (e.g., noun, verb, etc.) based on both its definition and context.
\item \emph{Syntactic Parsing:} the process of constructing a treelike structure of a given text that represents both POS tags of the words and the tags of syntactically related word groups (e.g., noun phrase).
\item \emph{Semantic Parsing:} the process of converting a given text into a formal knowledge representation that can be processed by software.
\item \emph{Semantic Role Labeling:} also known as shallow semantic parsing; the process of identifying constituents as the semantic arguments of each and every verb and determining their roles such as agent, instrument, and so on.
\item \emph{WordNet:} an English lexical database that arranges the words in an ontological representation based on relations such as synonymy, hypernymy, and so on \cite{fellbaum1998wordnet}.
\item \emph{FrameNet:} an English lexical database containing manually annotated sentences for semantic role labeling \cite{baker1998berkeley}.
\item \emph{ConceptNet:} a semantic network of common-sense knowledge \cite{liu2004conceptnet}.
\end{itemize}

\section{Text-to-Picture}
Text-to-picture conversion is probably the simplest method for visualizing natural language descriptions \cite{joshi2006story,zhu2007text,agrawal2011enriching,joshi2004story}. It treats the problem of mapping natural language descriptions to a visual representation as a data-driven image retrieval and ranking problem and tries to solve it using the foundations of commercial Web-based image search engines. In this approach, descriptive terms or constituents that represent the main concepts of the text are extracted using text-mining techniques such as bag-of-words, named entities, and N-gram models. It is assumed that an annotated dataset of images is available. In case of automatic annotation, it is common to collect a repository of multi–modal information containing both images and text and then to use the co-occurring text around images to annotate them. In Web-based image retrieval systems, this process is carried out by exploiting the surrounding text of the images and the text appearing within HTML tags. The extracted text is then tokenized and a subset of terms is  selected and scored to determine the weighted annotations of the corresponding images \cite{srinivasarao2012web,chen2012annotation,chiang2013interactive,gong2006web,kilincc2011expansion}. The extracted concepts are then matched against the image annotations and a subset of images are retrieved and ranked for a given concept based on some predefined similarity measures. Finally, for each concept, the retrieved images with the highest rank are illustrated in the same order that their corresponding concepts appear in the text.

This approach inherits the solid theoretical foundations of search engines. Also, because of exploiting statistical information retrieval rather than natural language understanding, the text-to-picture conversion approach is computationally efficient \cite{zhang2010understanding}. However, it does not result in expected visualization due to three main reasons: (1) It cannot capture the deep semantics embedded within the natural language descriptions, (2) the visualization is restricted to available images, and (3) it cannot interpolate the in–between visual information. This approach is not the main focus of this survey and hence only two systems are discussed.

\subsection{Story Picturing Engine}
The story picturing engine \cite{joshi2004story,joshi2006story} addresses the mapping process of a given textual story to a set of representative pictures by focusing on \quotes{ \emph{quantifying image importance in a pool of images.}} This system receives input stories such as \quotes{\emph{Vermont is mostly a rural state. The countryside has the cozy feeling of a place which \ldots}} \cite{joshi2006story} and ranks the related and available images accordingly as the output.

This system is a pipeline of three processes as follows. First, the descriptor keywords are extracted from the story. For this purpose, the stop-words are eliminated using a manually crafted dictionary and then a subset of the remaining words is selected based on a combination of bag-of-words model and named-entity recognition. The utilized bag-of-words model uses WordNet to determine the polysemy count — the number of senses
242 of a given word — of the words. Among them, nouns, adjectives, adverbs, and verbs with a low polysemy count (i.e., less ambiguity) are selected as descriptor keywords. A naive named-entity recognizer is used to extract the proper nouns such as names of places and people based on the beginning  letter of the words. Those images that contain at least one keyword and one named entity are retrieved from a local annotated image database. The next step in the pipeline is to estimate the similarity between pairs of images based on their visual and lexical features, which is calculated based on a linear combination of Integrated Region Matching (IRM) distance \cite{wang2001simplicity} and WordNet hierarchy. Finally, the images are ranked based on a mutual reinforcement method and top k ranked images are retrieved. This  system is basically an image search engine that gets a given description as a query and retrieves and ranks the related images. Despite the good accuracy and performance of the story picturing engine, it only retrieves one picture for a given story and ignores many aspects such as the temporal or spatial relations.

\subsection{Text-to-Picture Synthesis System}
The goal of this system is to augment the human-human and human-computer communications by adding a visual modality to the natural language channel \cite{zhu2007text}. In contrast to the story picturing engine, this system associates a different picture to each extracted key phrase and presents the story as a sequence of related pictures. It treats the text-to-picture conversion problem as an optimization process that tries to optimize the likelihood of the extracted key phrases, images, and placement given the input description. To extract the key phrases, the system first eliminates the stop words and then uses a POS-tagger to extract the nouns, proper nouns, and adjectives. These words are then fed to a logistic regression model to decide the probability of their picturability based on Google Web hit counts and image hit counts. Then, the TextRank algorithm \cite{mihalcea2004textrank} is applied to the computed probabilities and the top 20 keywords are selected and used to form the key phrases. The image selection
process is based on matching the extracted key phrases against the image annotations. If the matching is a success, then the matched images are retrieved. Otherwise, an image segmentation and clustering algorithm is applied to find an image that is more likely associated with the query key phrase. Ultimately, the retrieved pictures are positioned based on three constraints including minimum overlap, centrality of important pictures, and closeness of the pictures regarding the closeness of their associated key phrases. Despite the superiority of this system over story picturing engine, it still inherits the drawbacks of text-to-picture systems and results in stilted visualizations.

\section{Text-to-Scene}
One possible way to improve the visualization is to directly create the scene rather than showing representative pictures. This approach, known as text-to-scene conversion paradigm, lets the system elaborate on background, layout, lighting, objects, poses, relative sizes, spatial relations, and other features that cannot be addressed using text to-picture conversion systems \cite{coyne2001wordseye}. In a text-to-scene conversion system, words with specific POS tags carry more visual information than others. Noun and proper-noun POS tags are usually associated with objects, agents, and places and the words with these tags can be exploited to retrieve three-dimensional (3D) models from model repositories. An adjective POS tag is usually associated with a set of object features and the words with this tag are utilized to alter the object attributes such as color, relative size, etc. A preposition POS tag is mostly associated with spatial relations, and verbs usually determine actions and poses of articulated models such as an avatar pointing to an object.

The text-to-scene approach can generate elaborated and unified visualization of given descriptions within a single static scene, which is a far more coherent realization in comparison with the text-to-picture approach. Nevertheless, it faces the challenges mentioned in Section 1.1 such as designing NLU and KR components. Also, because the generated scene is static, it can address neither the dynamics nor the temporal relations and is only useful for visualizing a single episode. In this section, we will overview seven text-to-scene conversion systems.

\subsection{NALIG}
Natural Language Driven Image Generation (NALIG) is one of the early projects on generating static 2D scenes from natural language descriptions \cite{adorni1983natural,adorni1984natural}. It uses a very restricted form of input language that is basically a simple regular expression. The main focus of NALIG is to investigate the relationship between the spatial information and the prepositions in Italian phrases. The accepted form of phrases in this system is as follows:

\begin{center}
$[subject][preposition][object]$
\end{center}

Using this regular expression, NALIG can understand inputs such as \emph{the book is on the table.} It can also handle ambiguities within the phrases and infer simple implicit spatial arrangements using taxonomical rules such as Object X supports object Y that define the relations between the existing objects. These rules are defined based on state conditions, containment constraints, structural constraints, and supporting rules. For example, given an input such as \emph{a branch on the roof} the system can infer that \emph{a tree near the house having a branch on the roof.} In addition to spatial arrangements, NALIG also utilizes statics to infer how an object can support another object based on a physical equilibrium. All in all, NALIG is a very restricted system that does not support user interactions, flexible inputs, or 3D spatial relations.

\subsection{PUT}
The PUT language-based placement system \cite{clay1996put} is a rule-based spatial-manipulation system inspired by cognitive linguistics. It generates static scenes through direct manipulation of spatial arrangements of rigid objects using a restricted subset of the natural language. Using this restricted grammar, PUT is able to put 3D objects on top of each other or hang a 2D object on a 3D one. It can also disambiguate simple spatial relations such as \emph{on the wall} and \emph{on the floor.} This system consists of a simple parser implemented in C++ designed for its restricted input language and a rendering engine to visualize the static 3D environment. The syntax of the restricted language is in the form of a regular expression as follows.

\begin{center}
$[V][TR][[P][LM]]^+$
\end{center}

$V$ denotes the placement verb that specifies the type of positioning of the object being manipulated. The system only defines two placement verbs including put and hang. $TR$ represents the object being placed, whereas $LM$ represents the reference object. The system contains a set of 2D objects, such as walls and rugs, and a set of 3D objects, such as tables and lamps, that are already included in the virtual world. Hence, the user is limited to a set of pre-existing objects. $P$ is preposition and indicates the spatial relation between $TR$ and $LM$. Ten different groups of spatial relations such as above/below, left/right, and on are defined in this system. The Kleene plus operator lets the system handle compound spatial relations with a set of reference objects. As an example, an input command such as \emph{Put the box on the floor in front of the picture under the lamp} is decomposed to:

\begin{center}
$[put]_V[the box]_TR[on the floor]_P–LM[in front of the picture]_P–LM[under the lamp]_P–LM$
\end{center}

which consists of three consecutive prepositions constructing a chain of spatial relations.

In this system, objects are annotated by their names, which are used to match the geometric information of 3D models with their corresponding objects. The placement is carried out using axis-aligned bounding boxes of the objects to facilitate determining the surface and interiors of the objects. A simple failure handling mechanism is also used to handle the non-existent locations. In comparison with NALIG, PUT has a few advantages, such as more flexible allowed inputs, spatial arrangements, 3D object repository, and object manipulations. Nevertheless, it inherits a few disadvantages of NALIG, such as being restricted in terms of input language and interactions. Also, it only focuses on spatial relations and ignores  other clues within the scene description that can be used to infer the implicit knowledge.

\subsection{WordsEye}
WordsEye \cite{coyne2001wordseye} is designed to generate 3D scenes containing environment, objects, characters, attributes, poses, kinematics, and spatial relations. The system input is a set of textual descriptions that can include information about actions, spatial relations, and object attributes. The system consists of two main components, including a linguistic analyzer and a scene depicter. The linguistic analyzer is equipped with a POS tagger and a statistical parser. In the early version, an analyzer was implemented in common Lisp and, later, MICA parser \cite{bangalore2009mica} was exploited as well. It parses the input text and constructs a dependency structure that represents the dependencies among the words to facilitate the semantic analysis. This structure is then utilized to construct a semantic representation in which objects, actions, and relations are represented in terms of semantic frames \cite{coyne2010frame}. The words with noun POS tags are associated with 3D objects and their associated hyponyms (i.e., words with an is-a semantic relation) and hypernyms (i.e., words with inverse semantic relation of hyponymy) are acquired using WordNet. The spatial relations are captured using a set of predefined spatial patterns based on the dependency structure. The words with a verb POS tag are associated with a set of parametrized functions that indicate the effects of the verbs.

In a recent development, lexical knowledge extracted from WordNet and FrameNet are semi-manually refined to construct a Scenario-Based Lexical Knowledge Resource (SBLR), which is essentially a lexical knowledgebase tailored to represent the lexical and common-sense knowledge for text-to-scene conversion purposes \cite{coyne2010spatial}. The knowledge in SBLR is represented by VigNet \cite{coyne2011vignet,coyne2012annotation} which is an extension of FrameNet and consists of a set of intermediate frames called Vignettes that bridge the semantic gap between the semantic frames of FrameNet and the low-level graphical frames. VigNet also contains implicit knowledge of a restricted set of environments such as a kitchen. This knowledge is a remedy for missing common-sense facts within the natural language descriptions and is acquired through manual descriptions of the pictures gathered from Amazon Mechanical Turk (AMT) \cite{fort2011amazon}-an online crowd-sourcing framework for data collection using Human Intelligence Task (HITs). The collected corpus is processed using naive text processing techniques to populate the VigNet with extracted Vignettes \cite{rouhizadeh2010data,rouhizadeh2011collecting1,rouhizadeh2011collecting2}.

The Depiction module converts a set of semantic frames into a set of low-level graphical specifications. For this purpose, it uses a set of depiction rules to convert the objects, actions, relations, and attributes from the extracted semantic representation to their realizable visual counterpart. The geometric information of the objects is manually tagged and attached to the 3D models. This component also employs a set of transduction rules to solve the implicit and conflicting constraints while positioning the objects in the scene in an incremental manner. As soon as the layout is completed, the static scene is rendered using OpenGL.

WordsEye relies on its huge off-line rule-base and data repositories. Its semantic database consists of 15,000 nouns and 2,300 verbs, whereas its visual database consists of 2,200 3D models and 10,000 images. Different features of these models, such as geometric shape, type, flexibility, embeddability, and so on, are manually annotated. As an instance, all objects with a long thin vertical base are annotated as $stem$. WordsEye also contains a large set of rules including spatial rules, depiction rules, and transduction rules. For example, it contains three rules for $kicking$ action whose firing strengths are evaluated based on the type of object to be kicked. WordsEye has been utilized by a few thousand on-line users to create 15,000 static scenes. Although it has achieved a good degree of success, the allowed input language for describing the scenes is stilted \cite{coyne2010frame}. Another problem is that WordsEye is not interactive and does not exploit the user's feedbacks. Also, there is no interface provided for adding new knowledge or rules to the system and one would require hard-wiring them into the system.

\subsection{AVDT}
Automatic Visualization of Descriptive Texts (AVDT) \cite{spika2011avdt} generates static 3D scenes from descriptive text by emphasizing the spatial relations and the naturalness of the generated scene. It consists of two  layers, including an automatic scene graph generation layer and an object arranging layer. The former is responsible for processing and extracting the embedded information within the text and generating the scene graph from this information. It utilizes GATE \cite{cunningham2002framework}-an open-source text processing tool as a pre-processor for tasks such as lemmatizing the nouns, POS tagging, and generating dependency structures. It refines the pre-processed text by segmenting it into a few blocks based on punctuation marks and the coordination conjunctions and then assigns meta-data to the prepositions and the nouns within the text and ignores the rest. The prepositions are grouped according to their semantic similarities. For example, under, below, and beneath prepositions are classified into the under group. The meta-data contain the word role (i.e., preposition, dependent, or supporter), its position in the text, its quantity, and the corresponding
3D model. In the next step, a directed graph is constructed in which a node represents an object and an edge represents a preposition. The constructed graph is then pruned by merging the redundant nodes. This graph provides an efficient data structure for traversing the spatial relations.

The object arranging layer uses the described graph to render the scene. It assigns an axis-aligned bounding box for each object and applies distance and rotation heuristics for standardizing the scales and orientations of the dependent and supporting objects. The rotation heuristic ensures that a dependent object faces its supporter. It also applies a little randomness to achieve an untidy appearance. These heuristics result in a more natural appearance of the scene. AVDT focuses on the naturalness of the generated layout using manually crafted heuristics and proper analysis of spatial relations and, hence, results in more natural-looking scenes in comparison with WordsEye. Also contrary to WordsEye, AVDT can deal with linguistic cycles and allows more comfortable inputs. As an example, it is shown that WordsEye fails to visualize a sentence such as \emph{On the table is a vase} whereas AVDT can successfully realize it. In general, AVDT inherits the problems of WordsEye, such as stilted input, lack of interactivity, and relying on hand-crafted rules.

\subsection{System Developed at Stanford University}
Contrary to previous systems, the system developed at Stanford University \cite{chang2014interactive,chang2014learning,chang2014semantic} infers the implicit relations and partially supports interactive scene manipulation and active learning. In this system, a scene template is generated from an input text and converted to a geometric graph that is then utilized to render a static scene. The scene template constructed from the input text using Stanford CoreNLP language processing tool \cite{manning2014stanford} is a graph with objects as its vertices and relations as its edges. The objects are recognized by detecting nouns that are considered as visualizable according to WordNet. The words with an adjective tag within the noun phrases are extracted to identify the attributes of the objects. The spatial relations are extracted using a set of pre-defined patterns. Natural language descriptions usually do not contain common-sense facts about the spatial arrangements. To alleviate this challenge, the system uses conditional probability to model the object occurrences and hierarchy priors and exploits Bayes's rule to infer the implicit spatial arrangements. The inferred knowledge is then inserted into the scene template graph. For example, for a sample input text, \emph{put the cake on the table} it can infer to put the cake on a plate and put the plate on the table.

The geometric graph contains a set of 3D model instances that correspond to the objects within the scene template and their associated spatial arrangements. This graph is used directly to render the static 3D scene. The system also supports interactive scene manipulation and active learning. The user can add new objects to defined positions and remove the existing objects. She can also select an object within the scene and annotate it. The system modifies its probabilistic model of support hierarchy by observing how users design the scenes. For example, if a user asks the system to put a cup on the table, the system increases the probability of co-occurrence of the cup and the table and the probability of table supporting a cup. This system surpasses previous text-to-scene conversion systems in terms of adaptive behavior and interactivity. However, in terms of language understanding and richness of the model repository, WordsEye and AVDT outperform this system.

\subsection{Systems That Learn Visual Clues}

The systems mentioned so far only utilize textual clues either as a set of pre-defined rules or a set of models learned from corpora. A new trend in text-to-scene conversion systems is learning the associations between both textual and visual clues. These research works follow the approach used by text-to-picture conversion systems by focusing on learning the visual features from available image database and extracting associations between visual and textual features to automate the visualization process.However, contrary to the text-to-picture conversion systems, they use associations to position the objects within a static scene rather than selecting representative pictures.

AttribIt \cite{chaudhuri2013attribit} is designed to help the users create visual content using subjective attributes such as \emph{dangerous airplane}. This system provides the user with a set of 3D parts of a model of interest and helps her with assembling those components to construct a plausible model. It exploits AMT crowd-sourcing and presents the volunteers with a set of 3D models of different parts of objects such as airplane wings and asks them to compare each and every pair of models using adjectives. It then ranks the associations between the components and gathered attributes using the Support Vector Machine (SVM) classifier. The learned model is used along with a GUI to directly capture the semantic attributes and to provide the user with corresponding parts of the model.

A promising data-driven system developed at Microsoft Research Center is introduced in \cite{zitnick2013learning}. This system learns the visual features from abstract scenes, extracts the semantic information from corresponding corpus, and learns the associations between extracted visual features and semantics to generate  new scenes based on a set of unseen natural language descriptions. Similar to AttribIt, this system utilizes AMT for gathering a training dataset. It exploits Conditional Random Field (CRF) \cite{shotton2009textonboost} to extract objects and their occurrences, attributes, and positions. After extracting visual features, the system extracts semantics in the form of predicate tuples using semantic role analysis \cite{quirk2012msr}. For scene generation, the system learns the associations between the predicate tuples and the visual features based on their co-occurrences using highest mutual information and then uses these associations to generate a new scene. This system is purely data driven and learns how to generate static 2D scenes by observing available scenes and corresponding descriptors. However, it does not support online learning. The authors have only used their system on a simple 2D scenario of children's playground. Therefore, it is not clear whether their approach can generate satisfactory scenes in 3D scenarios as well. The system also lacks a strong semantic analysis for capturing more general dependencies.

\section{Text-to-Animation}
The text-to-animation paradigm adds dynamics to static scenes and realizes temporal relations as an extra layer towards the naturalness of the generated visualization. In this paradigm, in addition to linguistic analysis performed by the text-to-scene conversion systems, the visual verbs within the text are captured, parametrized, and then grounded to a set of virtual actions and manipulations within the digital world. The action parametrization is a big challenge in the case of general-domain systems and requires inference of knowledge about trajectories, targets, intermediate actions, and so on. Moreover, in a text-to-animation conversion system, the constraint network is expanded to capture the spatio-temporal constraints rather than just static spatial constraints. In other words, the objects may enter or exit the scene and the spatial relations among  them may vary as the simulation time proceeds. This approach can visualize the imaginations in a more natural way than the two aforementioned approaches. In this section, we will overview 17 text-to-animation conversion systems. It is noteworthy that we classify the systems in which user controls embodied agents by natural language commands as text-to-animation systems. The reason is that similar to conventional text-to-animation conversion systems, these systems manipulate the environment based on some verbal descriptions or commands as well. The only difference is that, in these systems, the user manipulates the world through the embodied agents rather than directly manipulating the objects.

\subsection{SHRLDU}
SHRLDU, developed by Winograd \cite{winograd1971procedures} at Massachusetts Institute of Technology, was one of the pioneer systems that integrated AI into computer graphics. It was also one of the early systems that used deep semantic parsing. SHRLDU consists of a simulated robotic manipulator equipped with an intelligent controller that operates within a virtual toy world. The world contains a few blocks with different shapes (e.g., cube, pyramid, etc.), sizes, and colors. The robotic arm can perform three actions on these blocks including (1)moving a block to a location, (2) grasping a block, and (3) ungrasping a block. The robot manipulates the environment according to a restricted set of given natural language commands. SHRLDU is implemented in the Micro-Planner and Lisp programming languages. Its architecture consists of four modules, including a language analyzer, planner, dialogue manager, and graphical engine.

The language analyzer operates based on a systematic grammar view of the language. It validates the syntactic analysis with semantic clues acquired from the environment through the parsing process. As an instance, for an ambiguous command such as \emph{Put the red pyramid on the block in the box} it first recognizes \emph{the red pyramid} as a possible noun phrase and then checks the world model to determine whether a unique red pyramid exists. Based on this observation, it then decides whether \emph{on the block} is part of the noun phrase. The planner component is used to plan a sequence of feasible actions to reach a goal state defined via input command. It utilizes a backward chaining algorithm considering the preconditions of actions to plan the sequence of manipulations. For example, given that in the world model \emph{the red block is on top of the blue block} and the user asks to \emph{put the pyramid on the blue block,} the robot first has to grasp the red block, move it to a random location, ungrasp it, grasp the pyramid, move it to the top of blue block, and, finally, ungrasp it. An interesting feature of SHRLDU is its dialogue manager, which enables it to answer simple queries regarding
the world configuration and history of actions it has taken. It also can ask for command
555 clarification in case of ambiguities in the input command and acknowledge the accomplishment of the tasks. Despite its restricted grammar, operational environment, and
naive dialogue manager, SHRLDU has inspired many other systems.

\subsection{PAR}
Parameterized Action Representation (PAR) developed at the University of Pennsylvania, is a framework for controlling virtual humans using natural language commands in a context-sensitive fashion \cite{badler1999animation,badler2001Parameterized,bindiganavale2000dynamically}. The main focus of PAR is to develop a comprehensive knowledge representation scheme to reflect the input commands on agents' behaviors. The structure of PAR contains applicability condition, start, results, participants, semantics, path, purpose, termination, duration, manner, sub-actions, parent action, previous action, concurrent action, and next action \cite{badler2001Parameterized}. 

In this representation, the applicability condition is a Boolean expression that indicates the feasibility of an action for a given agent. The start and result indicate the states and time stamps of a given action performed by the agent and the beginning and termination of that action. The participants refer to the agent that is executing the current PAR and the passive objects that are related to that action. The semantics include a set of Boolean pre-conditions and post-conditions of an action that must be satisfied to let the agent perform that action. They also embed the motion and the force of the action that should be applied. The path denotes the start and end points, direction, and distance of a motion. The purpose determines whether the current action should satisfy a set of conditions or trigger another action and the termination determines the condition for terminating the current action. PAR structure also contains a set of pointers to other PARs, including parent, next, previous, and concurrent actions.

The execution architecture of PAR implemented in C++ and Python is a reactive framework designed to handle the PAR representation. The architecture consists of five components, including language converter, database manager, execution engine, agent process, and visualizer. The language converter parses the input commands using an XTAG parser \cite{paroubek1992xtag} and uses a naive string matching algorithm to find the corresponding 3D objects and agents from the database through the database manager. It also captures the verbs and the adjectives in order to construct the corresponding PAR representation of the given verbal command. The execution engine synchronizes the actions using its universal clock and passes the received PAR structures to the corresponding agent processes. It also controls the visualizer to update the environment. Each active agent within the virtual world is assigned with an agent process that handles a queue of PARs (i.e., Pat–Net data structure) to be executed \cite{badler1993simulating}. The visualizer uses OpenGL to render the virtual world and its inhabitants are based on the received commands from the execution engine. The PAR architecture relies on shallow parsing rather than attempting to capture the deep semantics. It also does not support the deliberative planning that is essential for generating plans for complicated goals. The lack of interactivity is another drawback of this architecture.

\subsection{Carsim}
Carsim \cite{dupuy2001generating,aakerberg2003carsim,johansson2004carsim} is a domain-specific system developed for generating simple animations of car accidents based on a set of Swedish accident reports collected from news articles, narratives from victims, and official transcriptions from officers. It consists of two main modules including information extraction module and visualization module. The information extraction module analyzes the input text and converts it to a triplet representation <S,R,C> in which S denotes scene objects such as weather, R represents a set of road objects such as cars, and C is a set of collisions that happened in the accident. This module utilizes the Granska POS tagger \cite{carlberger1999implementing} for tagging the input text and uses a small lexicon and a few regular expressions to extract the named entities, such as street names. It also exploits a local dictionary extracted from WordNet to discover the action verbs. A light domain-specific ontology combined with a classifier that is trained using a small set of example reports are employed to extract the events from textual description of the accidents. The ontology is also utilized to solve the co-references.

The visualization module utilizes an animation planner and a graphical engine to render the planned animation. The animation planner exploits a naive greedy algorithm to plan the animation considering the constraints, initial positions, initial directions, and the trajectories. The planning algorithm does not support backtracking and thus cannot find the optimal plans. The constraints are addressed using a small set of spatial and temporal rules. The initial direction and position are inferred directly from the input report and then propagated to those objects whose initial condition is not explicitly mentioned in the report. The trajectories are acquired using the Iterative deepening $A^*$ $(IDA^*)$ algorithm. Carsim is a good example of a practical text-to-animation conversion system that mostly focuses on the practical aspects rather than theoretical arguments. It has shown a fair degree of success in its limited domain. Yet it lacks a solid mechanism to harvest the information from user interactions and feedbacks. It also does not contain a strong object repository or lexical resources.

\subsection{ScriptViz}
ScriptViz \cite{liu2006script} aims to replace the manual storyboard drawing with automatic dynamic scene generation in the motion-picture production process. It is capable of analyzing the screenplays written in well-formed sentences (i.e., grammatically correct and not ambiguous) and animating the corresponding objects, agents, and actions. The system consists of three interacting modules, including a language understanding module, a high-level planner, and a scene generator. The language understanding module uses Apple Pie parser \cite{sekine1998corpus} to derive the syntactical structure of the input text. It separates the clauses based on the conjunctions, extracts the actions from verbs, and recognizes the objects from proper nouns. The verb and proper noun are matched against the actions and the objects using a naive binary matching mechanism, respectively. 

The high-level planning module generates action plans based on the information received from the language understanding module. The planning process is completed within four consecutive phases. First, an offline plan outline is extracted from a plan database in respect to the objects and actions detected in the input script. The states of the objects and agents are then collected from the virtual environment. This information is used to decide the feasibility of actions according to the current configuration of the environment. In case of a feasible action, parameters of the offline plan are set and the result is represented using PAR structure \cite{badler2001Parameterized}. The scene generator assigns the resulted PAR to the corresponding agent, updates the states, and renders the scene in real time. ScriptViz is implemented in Java and uses OpenGL as its graphical engine. It does not support interactive modification of the generated animation and does not embed any lexical or common-sense resources. Also, it has a very limited model repository and scene layout options. These limitations result in weak visualizations of given scripts. Furthermore, it is not clear what kind of actions the agents can perform as the treatment of articulated bodies is not discussed in this work.

\subsection{CONFUCIS}
CONFUCIS \cite{ma2006automatic} is a multi-modal text-to-animation conversion system that can generate animation from a single input sentence containing an action verb and synchronize it with speech. It is basically a narrator system developed for animating human characters with a peripheral narrator agent for storytelling of the actions. CONFUCIS can address the temporal relations between the actions performed by the virtual humans. It utilizes H-Anim standard for modeling and animating the virtual humans. It supports lip synchronization, facial expressions, and parallel animation of the upper and the lower body of human models \cite{ma2004visual}.

CONFUCIS consists of a knowledgebase, language processor, media allocator, animation engine, text-to-speech engine, narrator, and synchronizer. The knowledgebase contains a lexicon, a parser, and a visual database. The visual database contains a very limited set of 3D models and action animations. The language processor uses a Connexor functional-dependency Grammar parser
\cite{tapanainen1997non}, WordNet, and a lexical conceptual structure database \cite{ma2006virtual} to parse the input sentence and capture the semantics it carries. The media allocator exploits the acquired semantics to generate an XML representation of three modalities, including animation, speech, and narration. The animation engine uses generated XML and the visual database to generate animation. The text-to-speech and the narrator modules also use the XML to generate speech and initialize the narrator agent, respectively. Finally, the synchronizer integrates these modalities into a VRML file that is later used to render the animation.

One of the main challenges of a text-to-animation conversion system is defining a set of sub-actions that can result in a high-level action. In a hypothetical scenario, assume that the input sentence is \emph{John hits Paul with a bottle and John is in a distance of 2m from Paul and there is a bottle on a table that is in a distance of 1m from John.} To realize this input sentence with a plausible animation, the system should exploit a planner to schedule a set of intermediate actions such as \emph{John walks toward the table, picks up the bottle, walks toward Paul, and hits him with the bottle}. CONFUCIUS addresses this challenge by using hand-crafted sub-actions that in turn restrict the animation to a few predefined actions (i.e., less than 20 visual verbs). Also, due to limited number of sentences (i.e., one sentence) in each input and the restricted format of the input sentences (i.e., one action verb per sentence), the user is restricted in expressing the intended description. CONFUCIUS is not interactive in a sense that it does not let the user modify the generated animation.

\subsection{Scene Maker}
SceneMaker \cite{hanser2009scenemaker,hanser2010scenemaker} is a collaborative and multi-modal system designed for pre-visualizing the scenes of given scripts to facilitate the movie production process. This system is a successor of the CONFUCIS system and exploits its underlying language processing and multi-modal animation generation tools. SceneMaker expands CONFUCIS by adding common-sense knowledge for genre specification, emotional expressions, and capturing emotions from the scripts. Users can edit the generated animation online via mobile devices.

SceneMaker consists of two layers, including a user interface that can run on a PC or a mobile device and a scene production layer running on a server. The user interface receives a screenplay from the user and provides her with a 3D animation of the script and a scene editor to edit the generated animation. The scene production layer contains three components operating in a serial manner, including an understanding module, a reasoning module, and a visualization module. The understanding module performs text analysis using the CONFUCIS platform. The reasoning module uses WordNet-Affect \cite{strapparava2004wordnet}-an extension to WordNet and ConceptNet to interpret the context, manage emotions, and plan the actions. The visualization module fetches the corresponding 3D models and music from the database, generates speech, and sets the camera and lighting configuration. Despite adding interactions and alleviating input restrictions, SceneMaker inherits the flaws of CONFUCIS in terms of action definition. It is noteworthy that we could not find any snapshots of the resulting animation in the published articles.

\subsection{System Developed at Kyushu Institute of Technology}
This system is designed to generate motion for virtual agents using a set of motion animations stored within a motion database \cite{oshita2009generating,oshita2010generating}. To carry out this task, it captures pre-defined action verbs including intransitive (no target object), transitive (one target object), and ditransitive (two target objects) verbs from the input using a local dictionary. The system exploits motion frames—an extension to case frames focusing on semantic valence \cite{fillmore1967case}-as its knowledge representation scheme, which consists of an agent, a motion, an instrument, a target, a contact position, a direction, an initial posture, and a set of adverbs to modify the motion. It is assumed that the characters, objects, and motion frames are manually predefined by the user. The workflow of this system is as follows. 

First, the input sentence is parsed using the Stanford CoreNLP tool and then a small set of rules (e.g., four rules for temporal constraints) and a dictionary are utilized to extract the query frames and the temporal constraints. The query frames are motion frames extracted from the input that are matched against the motion database. The temporal constraints determine whether two actions are serial or parallel. The system uses the extracted temporal constraints to create a rough schedule of the actions and searches the motion database to find the motion clips that match the query frames. The motion database consists of a set of manually annotated atomic motions represented in motion frames. These atomic actions can be combined to create more complex actions. The matching process is done in two consecutive steps. First, the query frame is matched against the motion frames in the database based on the actions and the agents. The retrieved candidates are then ranked using a weighted similarity measure based on the target, instrument, initial posture, and adverbs. The system can generate a set of intermediate motions such as locomotion and grabbing an instrument. Ultimately, the predefined scene information and the retrieved atomic motions are integrated in order to animate the motions.

This system relies on its offline motion database, which makes it difficult to handle unseen motions. It is also not clear how atomic motion clips are fused to generate compound motions. Another disadvantage of this system is its limited language processing capabilities. Last but not least, it imposes a high volume of workload on users by assuming that the characters, objects, and motion frames are manually predefined by the users.

\subsection{IVELL}
Intelligent Virtual Environment for Language Learning (IVELL) \cite{hassani2013architectural,hassani2016design} is a domain-specific multi-modal virtual reality system that consists of a few Embodied Conversational Agents (ECAs). It is designed to improve the speaking and listening skills of non-native users in English. IVELL implements a few scenarios, such as an airport and shopping mall, in which learners speak to domain-specific agents such as an immigration agent while manipulating the virtual world using a haptic robot. The agents can alter the difficulty level of the conversation by automatically evaluating the user's linguistic proficiency. Each agent consists of an abstract layer and an embodied layer.

The abstract layer consists of a language interpreter, user evaluator, fuzzy knowledgebase, haptic interpreter, language generator, and action coordinator. The language interpreter lemmatizes and parses the inputs using the OpenNLP tool and matches the results against deterministic finite automata to capture the user's intentions. The user evaluator uses a weighted model to score the user's proficiency. The knowledgebase is a light domain-specific fuzzy ontology that keeps knowledge about the predefined tasks. The haptic interpreter maps the low-level force and position vectors acquired from a haptic robot to high-level perceptions. The language generator generates a set of answers with different difficulty levels based on the knowledge extracted from the knowledgebase. The action coordinator synchronizes the graphical actions, haptic actions, and output speech whose score is the closest to the user's proficiency level. The embodied layer contains a speech recognizer, a text-to-speech engine, a haptic interface, and an avatar controller. The system is developed in C\#.Net and uses Autodesk 3DMax and 3DVIA Virtools to model and render the environment. Different modules within this system communicate synchronously through TCP/IP protocol which provides the system with distributed processing capabilities. 

IVELL is an interactive system that can adapt its interactions based on the user's proficiency level. It also utilizes a natural language generator to augment the interactions. Moreover, it asks user's help when it is not able to understand her utterance. Nevertheless, it uses a very limited approach to capture the semantics. Also similar to previous systems, it uses a set of limited and hardwired actions.

\subsection{Other Systems}
One of the early text-to-animation synthesis systems is the Story Driven Animation System (SDAS) introduced in \cite{takashima1987story}. This Japanese system consists of three modules, including story understanding, stage directing, and action generating modules implemented in the Prolog and Lisp programming languages. The system input is restricted to unambiguous text that can only contain sentences describing actions in a time sequence. The story understanding module performs syntactic and semantic analyses. However, the original article does not explain the applied techniques. It also uses an assumption-based reasoning that adds very simple implicit assertions about the story. The stage direction module exploits a few simple heuristics to position the actors and set the background based on the extracted information and generated assertions. The action generating module uses a set of model descriptions and motion descriptions. A very limited set of simple articulated figures and primitive joint motions are defined and combined to create a simple animation.

3DSV \cite{zeng20053d,zeng2005visual,zeng2007development} attempts to create an interactive interface for animating 3D stages of simple stories described in restricted sentences. The stage includes objects, their attributes, and simple spatial relations. The spatial relations are captured using a set of regular expressions and represented in XML format. 3DSV utilizes an XML-based knowledgebase to parametrize the extracted properties of the stage. The knowledgebase contains visual descriptions of the objects, attributes, and spatial relations. The information extracted from the input text and the knowledgebase are then integrated into an XML representation that is converted to the VRML format. The VRML is animated within a Java applet and lets the user manipulate the stage using mouse commands. Despite the simplicity and restricted nature of 3DSV, it provides the users with cross-platform functionalities.

Interactive e-Hon \cite{sumi2005automatic,sumi2006animated} is a Japanese multi-modal storytelling system designed for facilitating the interactions between parents and children by animating and explaining the difficult concepts in a simpler form using Web content. This system uses a Japanese morphological analyzer and a lexicon to extract time, space, weather, objects, and actions from the story. The extracted information is matched against two lookup tables, including a background table and an action table. The time, space, and weather are matched against the background table to provide the animation with the appropriate static background. The objects and the actions are matched against an action table that is used to retrieve a corresponding recorded animation from a database. This system mostly relies on lookup tables and binary matching algorithms, which severely limits its capability of semantic analyses.

A semi-automatic system developed at Rhodes University \cite{glass2008automating1,glass2008automating2} generates animations from given annotated fiction texts. The basic assumption in this system is that the characters, objects, environment configuration, spatial relations, and the character transitions in the text are annotated in a well-formed structure in advance. It uses the annotations of the characters and the objects to query a 3D model database. The system exploits a query expansion mechanism using WordNet to enhance the possibility of finding proper models. It also uses annotated spatial information to construct a spatio-temporal constraint network. It provides the users with an interface to alter the constraint network to increase the artistic aspects of the generated animation. The layout constraints are satisfied using an incremental greedy algorithm. The system is developed in Python, and the extracted models and the environment are rendered using Blender3D. This system lets the user modify the animation by manipulating the constraints and provides a robust model matching scheme using the query expansion mechanism. On the other hand, it requires annotated fiction texts as its input, which is a labor-intensive and tedious task. 

A data-driven system developed at the University of Melbourne \cite{ye2008towards}attempts to train a classifier to ground high-level verbs into a set of low-level graphical tasks. To carry out this task, it extracts verb features, collocation features, and semantic role features from the scripts. It also extracts binary spatial features from the virtual stage. These linguistic and visual features are then used to train a maximum entropy classifier \cite{ratnaparkhi1996maximum} to decide the next graphical action. Despite its interesting approach for co-training the semantic and stage features, it fails to provide a proper means of interaction.

Web2Animation \cite{shim2009web2animation} is a multi-modal pedagogical system that uses Web content related to recipes to create an online animation to teach the users how to cook. Converting the Web content to an animation is done within three steps, including extracting relevant text, capturing semantics, and animating actions. The relevant recipe information is located by traversing the HTML tags and analyzed using the Phoenix parser \cite{ward1989understanding}. The extracted instructions are mapped to a few actions and the captured ingredients are associated with a set of objects. A domain-specific ontology is utilized to match the actions with their graphical representation. The ingredients are also matched against their graphical models. Finally, a user-created screenplay is used to synchronize the animation with a monologue explaining the recipe. In this system, the user has to craft the screenplay, which interferes with its pedagogical purpose. Also, considering the noise-prone nature of Web content, it is not clear how well the system will behave in mining useful content.

Vist3D \cite{oddie2011applying,presland2010creating} is a domain-specific system for creating 3D animation of historical naval battles from narratives provided by the users. The system uses a manually populated ship specification database and a temporal database. The temporal database is populated by a narrative analyzer that extracts the time and date and $[Subject][Verb][Object]$ structures using a set of regular expressions. The retrieved information from these two databases is converted to VRML format. Vist3D is designed in a very restricted way. Similar to NALIG, it can only detect very simple syntactic structures and utilizes a very small dictionary.

A different approach that relies on service-oriented and multi-agent design methodology is proposed in \cite{bolano2011using}. It models the agents using NLP4INGENIAS \cite{moreno2009using}, which is a multi-agent system based on the INGENIAS framework \cite{pavon2003agent}. NLP4INGENIAS exploits natural language descriptions to model the agents and supports user-in-the-loop disambiguation of the descriptions. The acquired agent models are fed to Alice \cite{kelleher2007using}-a rapid prototyping environment for generating virtual environments to render the world and agents. This system approaches the text-to-animation conversion problem from a software engineering point of view. It exploits agile software development using existing platforms rather than struggling with theoretical sophistications. On the other hand, it is restricted to the limitations of its building blocks and cannot tailor them to meet its specific requirements. 

An adaptive animation generation system is introduced in Hassani and Lee \cite{hassani2015adaptive}. This system is a multi-agent and data-driven system that utilizes statistical Web content mining techniques for extracting the attribute values of objects such as relative sizes and velocities. The system consists of three interacting agents, including an information retrieval agent, a cognitive agent, and a language processing agent. The cognitive agent contains a knowledgebase and a planner to decide the actions. It also interacts with visualization interface in terms of high-level visual operations and perceptions. The authors mostly focus on the information retrieval agent and do not elaborate on the language processing agent. The reported accuracy of the retrieved results is promising. However, the results are only provided for simulating the solar system and it is not clear whether it can generalize to other scenarios as well. The language processing agent employs a set of regular expressions for extracting the embedded information and query generation.

\section{Discussion}
We elaborated on 26 systems including 2 text-to-picture conversion systems, 7 text-to-scene conversion systems, and 17 text-to-animation conversion systems. The evolution of the text-to-picture conversion systems can be identified in two main directions. (1) The systems have evolved in terms of extracting text-image associations. The early systems only exploit associations between the text and image annotations. Later, these associations are augmented by fusing the visual features with the semantic features. (2) The systems also have evolved in terms of output. The early system provides the users with only one representative picture, whereas the successor system provides the users with a set of images ordered based on the temporal flow of the input descriptions. The future text-to-picture conversion systems can improve by exploiting better semantic processing, image processing, and association learning techniques. However, because they are limited to pictures, the results will not enhance dramatically in comparison to the current systems.

We investigate the evolution of the text-to-scene conversion systems in terms of five measures, including lexical flexibility, grammatical flexibility, action diversity, spatial diversity, and object diversity. The lexical and grammatical flexibility measures are related to the flexibility of the input language, whereas the other three measures determine the quality of the output. These measures are defined based on the Likert scale and have five distinct values including +2 (very high), +1 (high), 0 (medium), -1 (low), and -2 (very low).
The evolution timeline of the text-to-scene conversion systems is illustrated in Figure 1. 

\begin{figure}
  \hspace*{-0.8cm}	\includegraphics[scale=0.75]{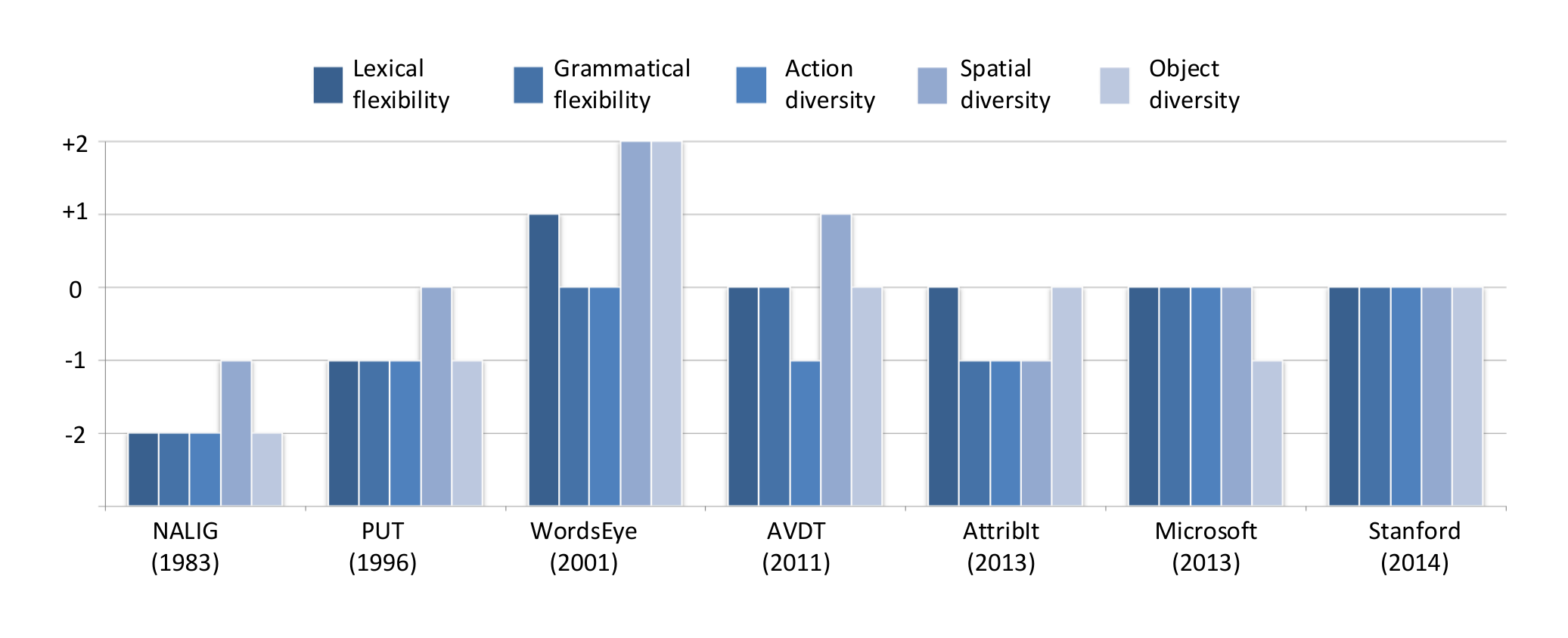}
  \caption{Evolution of the text-to-scene conversion systems in terms of lexical flexibility,  			grammatical flexibility, action diversity, spatial diversity, and object diversity.}
  \label{fig:Figure 1}
\end{figure}

As shown, the diversity of the input vocabulary and the flexibility of the input structure improve from NALIG to WordsEye and then are no longer enhanced. This trend represents the current technical difficulties in understanding natural language. The evolution of the action diversity follows a similar trend. Because of the large number of possible actions, it is not practical to craft all of them. On the other hand, learning actions and associating them with action verbs is a big challenge for the current machine vision techniques. In terms of spatial and object diversity, WordsEye has almost achieved a good performance by relying on its huge object database and large number of hand-crafted spatial rules. This is because the spatial relations are limited and can be hand crafted. An important observation is that the current data driven systems do not outperform the rule-based systems. This is probably because the
data-driven systems have been only used for feasibility studies, whereas a few rule based systems such as WordsEye are commercialized, which, in turn, provide them with the required resources for crafting as many rules as possible. Moreover, as far as the authors' knowledge is concerned, there is no big and useful dataset available for this purpose.

We investigate the evolution of the text-to-animation conversion systems with a similar approach by adding a temporal diversity measure to the measures used to evaluate the text-to-scene conversion systems. The evolution timeline is illustrated in Figure 2.

\begin{figure}
  \hspace*{-0.8cm}	\includegraphics[scale=0.75]{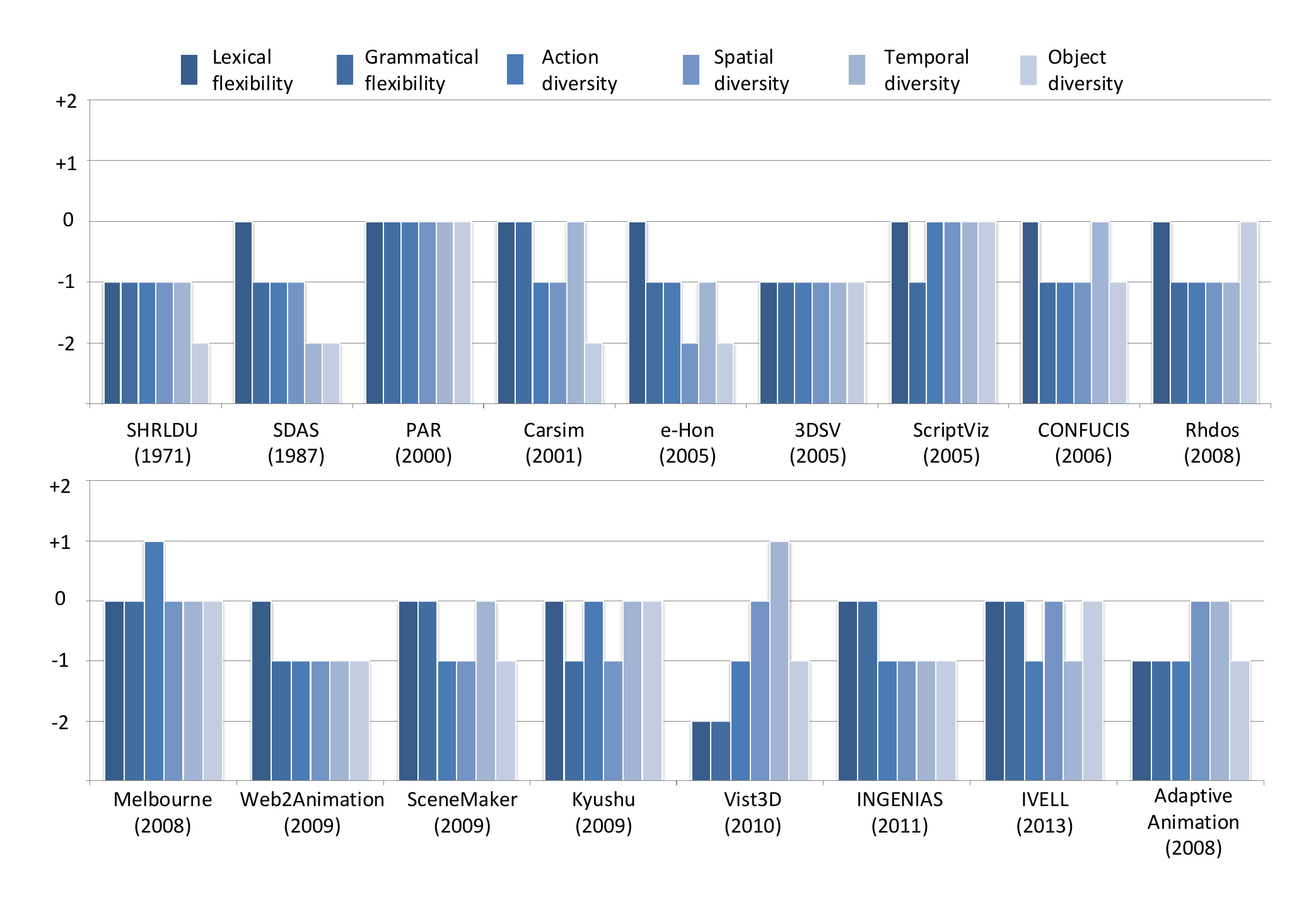}
  \caption{Evolution of the text-to-animation conversion systems in terms of lexical 		  	flexibility, grammatical flexibility, action diversity, spatial diversity, temporal diversity, and object diversity}
  \label{fig:Figure 2}
\end{figure}

Surprisingly, as shown in this figure, the trend indicates that the text-to-animation conversion systems have not improved much since SHRLDU. These systems can improve in terms of object diversity and spatial diversity using similar approaches taken by systems such as WordsEye. Also, because the temporal relations are more limited than spatial relations, this measure can be improved as well. Nevertheless, the text-to-animation conversion systems inherit the challenges related to the actions and the input language. We conclude that both the text-to-scene and the text-to-animation conversion systems will not significantly improve until the machine vision and language understanding methods are improved. Fortunately, the new advances in deep convolutional neural networks and long short-term memory neural networks are gradually enhancing these two areas of research, respectively.

We summarize the discussed systems in Table 1. The type of the system indicates whether it is a text-to-picture, text-to-scene, or a text-to-animation conversion system. The interactivity indicates whether it provides the user with means to manipulate the generated output, whereas the adaptive characteristic refers to the system's capability in extracting information that is not given to the system a priori. A system that uses data-driven techniques such as Web content mining, active learning, crowd-sourcing, association learning, and so on, is considered as adaptive. The interface type can be text, speech, or a pointer (i.e., mouse interaction). The system domain determines whether it is built for general or custom purposes. The syntactic and semantic analyses indicate the approaches that the system utilizes to analyze the text. Finally, the methodology and the knowledgebase determine the paradigm on which the system is built (i.e., data-driven, rule based, and multi-agent) and the exploited knowledge resource (e.g., lexicons), respectively.

\begin{table}
\caption{Characteristics of the existing language-based visualization systems}
\noindent
\hspace*{-0.5cm}\begin{tabular}{|l|l|l|l|l|l|l|l|}
\hline
System & Type & Interactive & Adaptive & Interface & Domain  \\ 
\hline
Story picturing engine & TTP & No & Yes & Typed text & General \\
Text-to-picture synthesizer & TTP & No & Yes & Typed text & General \\
PUT & TTS & No & No & Typed text & General \\
WordsEye & TTS & No & No & Typed text & General \\
AVDT & TTS & No & No & Typed text & General \\
Stanford University & TTS & Yes & Yes & Text + Pointer & General \\
NALIG & TTS & No & No & Typed text & General \\
AttribIt & TTS & Yes & Yes & Typed text & General \\
Microsoft & TTS & No & Yes & Typed text & General \\
University of Melbourne & TTS & No & Yes & Typed text & General \\
SHRLDU & TTA & Yes & No & Typed text & Specific \\
CONFUCIS & TTA & No & No & Typed text & General \\
SceneMaker & TTA & Yes & No & Typed text & General \\
ScriptViz & TTA & No & No & Typed text & General \\
Kyushu Institute & TTA & NO & No & Typed text & General \\
Carsim & TTA & No & No & Typed text & Specific \\
PAR & TTA & No & No & Typed text & General \\
IVELL & TTA & Yes & Yes & Speech+Pointer & Specific \\
SDAS & TTA & No & No & Typed text & General \\
Adaptive animation & TTA & Yes & Yes & Typed text & Specific \\
Interactive e-Hon & TTA & No & Yes & Typed text & General \\
Vist3D & TTA & No & No & Typed text & Specific \\
Web2Animation & TTA & No & Yes & Typed text & Specific \\
Rhodes University & TTA & Yes & Yes & Typed text & Specific \\
3DSV & TTA & Yes & No & Typed text & Specific \\
INGENIAS-based system & TTA & No & No & Typed text & General \\
\hline
\end{tabular}
\\
\noindent
\hspace*{-3.35cm}\begin{tabular}{|l|l|l|l|l|}
\hline
System & Syntactic Analysis & Semantic Analysis & Methodology & knowledgebase  \\ 
\hline
Story picturing engine & BOW & Association & Data-driven & WordNet \\
Text-to-picture synthesizer & POS-tagging &Association & Data-driven & None\\
PUT & Regular expression & None & Rule-based & None \\
WordsEye & Statistical parsing & Dependency & Rule-based & VigNet+WordNet\\
AVDT & Statistical parsing & Dependency & Rule-based & None\\
Stanford University & Statistical parsing & None & Data-driven & WordNet\\
NALIG & Regular expression & None & Rule-based & None\\
AttribIt & POS-tagging & Association & Data-driven & None\\
Microsoft & POS-tagging & Semantic role & Data-driven & None\\
University of Melbourne & Statistical parsing & Semantic role & Data-driven & None\\
SHRLDU & Dependency parsing & Semantic parsing & Rule-based & None\\
CONFUCIS & Dependency parsing & Lexical & Rule-based & WordNet+Conceptual database\\
SceneMaker & Dependency parsing & Lexical & Rule-based & WordNet-affect+ConceptNet\\
ScriptViz & Statistical parsing & None & Rule-based & None\\
Kyushu Institute & Statistical parsing & None & Rule-based & None\\
Carsim & POS-tagging & Ontology & Rule-based & WordNet+Ontology\\
PAR & Dependency parsing & None & Rule-based & None\\
IVELL & Statistical parsing & None & Multi-agent & Fuzzy ontology\\
SDAS & Unknown & Unknown & Rule-based & None\\
Adaptive animation & Regular expression & None & Data-driven & WordNet\\
Interactive e-Hon & Regular expression & None & Rule-based & None\\
Vist3D & Regular expression & None & Rule-based & None\\
Web2Animation & Statistical parsing & Ontology & Rule-based & Ontology\\
Rhodes University & Regular expression & None & Rule-based & WordNet\\
3DSV & Regular expression & None & Rule-based & XML-based knowledgebase\\
INGENIAS-based system & Statistical parsing & Active learning & Multi-agent & None\\
\hline
\end{tabular}
\end{table}

As shown in Table 1, in terms of system behavior only 30.7\% of the systems are interactive and only 42.3\% of them are adaptive. This behavioral information reveals a fundamental flaw in most of the research works carried out in this direction. A system designed for visualizing the natural language descriptions should be both interactive and adaptive. Considering the current technical challenges with designing a complete natural language understanding component, the system should harvest the relevance feedbacks and the modifications performed by the user to evolve in an incremental manner. The system can disambiguate the input text in collaboration with the user as well. Also considering the huge amount of common-sense information required for such system, it is not practical to gather the information manually. And, hence, data-driven methods (e.g., Web content mining, Corpus mining, etc.) and active learning (i.e., user-in-the-loop learning) should be integrated into these systems. Moreover, it is not possible to pre-determine all possible actions that a user may ask from the system.
These actions can range from a \emph{character shooting a gun} to a \emph{horse galloping on the hills.} To address this challenge, the system should be able to detect and capture the motions of the actions by learning the dynamics and features of the actions and re-target them to other agents. A potential but challenging approach would be using machine vision techniques to learn the actions from online annotated multi-media content such as YouTube. Both of these tasks (i.e., natural language understanding and learning actions) are currently a bottleneck for such systems. However, new developments in end-to-end learning paradigms, especially the combination of deep learning and reinforcement learning, has shown potentially promising results that can be applied to mitigate the mentioned challenges \cite{mnih2015human}.

In terms of interface type, only 7.7\% of the systems utilize mouse interactions (i.e.,
mentioned as a pointer in Table I) and only 3.8\% of them utilize speech. As suggested by the study reported in \cite{lee2014potential}, it is better to use a hybrid interface consisting of natural language and mouse interactions. Also, considering the current advances in speech recognition technology, it is simpler to use speech rather than typed text. In terms of the domain, 69.2\% of the systems are general-domain systems, whereas 30.8\% are designed as domain specific. Nevertheless, this percentage does not reflect the completeness of the systems. For example, even though Carsim is a domain-specific system, it outperforms some of the general-domain systems in terms of language processing and semantic analysis. Except for some cases, designing general-domain systems resulted in systems with more restrictions. This is because most of these systems ignore the adaptive and interactive behaviors and heavily rely on hard-wiring the rules. As summarized in Table I, 26.9\% of the systems are designed based on a data-driven approach, 7.7\% follow a multi-agent paradigm, and 65.4\% are rule-based systems. Surprisingly, 61.1\% of the general-domain systems are designed following a rule-based approach that in turn prevents them from supporting appropriate degrees of generalizability. A successful system should ignore neither a priori knowledge provided by the experts nor the chunks of knowledge that can be acquired from different online resources. Such a system also should be able to distribute the tasks among different agents to support cross-platform and service-oriented models. Therefore, a practical and general natural language visualizer requires the integration of a priori knowledge (rule based) with extracted knowledge through the process (data driven) while distributing the tasks among a few agents (multi-agent).

In terms of syntactic analyses, 26.9\% of the systems exploit regular expressions, 15.4\% of them utilize POS tagging, and 34.6\% of them employ syntactic parsing. Among general-domain systems, 61.1\% of them use syntactic parsing, 16.7\% of them exploit POS tagging resulting in loss of constituent information, and 16.7\% of them use regular expressions that essentially ignore the syntactic information. The latter two methods cannot provide proper syntactic analyses in comparison with parsing techniques. Therefore, 33.4\% of the general-domain systems suffer from this deficiency. In terms of the semantic analyses, 46.2\% of the systems rely on naive matching of the keywords, which is equivalent to ignoring the semantics. On the other hand, 53.8\% of the systems exploit some shallow semantic analyses. Shockingly, 33.3\% of the general-domain systems follow this approach. Furthermore, 15.4\% of the systems use knowledgebase and ontologies for semantic analysis whereas 7.7\% of them exploit user-in-the-loop semantic analysis. The rest of the systems rely on shallow semantic analysis as follows: 11.5\% association analysis, 7.7\% dependency analysis, and 7.7\% semantic role analysis.

Finally, in terms of using knowledgebase, lexicons, and ontologies, 57.7\% of the systems completely ignore these resources. This ratio is 72.2\% for general-domain systems. In other words, a great fraction of the general-domain systems that require common-sense knowledge are not equipped with any knowledge resources. This fact highlights another fundamental problem of the current systems: They simply ignore the knowledge resources and hence cannot infer in unpredicted situations. Among the systems utilizing some sort of knowledge resources, 63.6\% of them exploit the WordNet lexicon. 

All in all, we identify two main problems with the current systems. The first problem is associated with the current technical challenges such as natural language understanding, knowledge representation, common-sense knowledge, implicit knowledge, and action learning. To our surprise, the second problem is rooted in the fact that the current systems appreciate neither the available resources (e.g., lexicons and semantic networks) nor the available techniques (active learning, shallow semantic analysis, etc.) and, hence, do not meet the expected requirements for a natural language visualizer. The first problem possibly will be alleviated using end-to-end learning algorithms in the near future. Deep learning has shown promising results in machine vision, object recognition, speech recognition, and language modeling \cite{bengio2009learning}, and deep reinforcement learning has shown promising results in action learning \cite{mnih2015human}. The remedy for the second problem is to combine rule-based and data-driven paradigms to design adaptive and interactive systems that utilize available resources and techniques to its full extent.

\section{Conclusion}
In this article, we discussed the requirements and challenges for developing systems that are capable of visualizing descriptions expressed in a natural language. We reported 26 such systems and elaborated on the methodology; implementation; natural language processing aspects, including morphological, syntactic, and semantic analyses; knowledgebase; lexicons; AI components; computer graphics aspects, such as rendering and model repositories; and the pros and cons of these systems. We conclude that, in addition to the current technical challenges in natural language understanding, providing common-sense knowledge, inferring the implicit knowledge, action learning, and so on, most of the systems introduced in the literature appreciate neither the available resources (e.g., lexicons and semantic networks) nor the available techniques (active learning, shallow semantic analysis, etc.) and, hence, do not meet the expected requirements for a natural language visualizer. We predict that, by using end-to-end learning algorithms, the current challenges of developing these systems will be mitigated in the foreseeable future.

\bibliographystyle{plain}
\bibliography{NLsurvey_bib}
\end{document}